\documentclass[11pt]{article}

\usepackage[final]{acl}

\usepackage{times}
\usepackage{latexsym}

\usepackage[T1]{fontenc}
\usepackage[utf8]{inputenc}

\usepackage{microtype}
\usepackage{inconsolata}
\usepackage{graphicx}

\usepackage{amsmath,amssymb}
\usepackage{booktabs}
\usepackage{multirow}
\usepackage{enumitem}
\usepackage{url}
\usepackage{color}

\title{Topology-Enhanced Alignment for Large Language Models:\\
Trajectory Topology Loss and Topological Preference Optimization}

\author{
  Yurui Pan$^{1}$\thanks{Equal contribution.} \quad Ke Xu$^{2}$\footnotemark[1] \quad Bo Peng$^{3}$\thanks{Corresponding author.} \\
  $^1$School of Computing and Intelligent Innovation, Fudan University \\
  $^2$School of Economics and Management, Tongji University \\
  $^3$College of Information Technology, Shanghai Ocean University \\
  \texttt{yrpan24@m.fudan.edu.cn, kexu567@tongji.edu.cn, bpeng@shou.edu.cn} \\
}

\begin{document}
\maketitle

\begin{abstract}
Alignment of large language models (LLMs) typically relies on supervised fine-tuning (SFT) and reinforcement learning from human feedback (RLHF), or more recently direct preference optimization (DPO).  
However, existing objectives largely ignore the global geometry and topology of the representation space: they operate on local token-level likelihoods or scalar preference scores, and do not explicitly constrain how hidden states move from a user prompt to an answer.

We view generation as tracing a \emph{semantic trajectory} in hidden space, and propose a topology-enhanced alignment framework that regularizes these trajectories using $0$-dimensional persistent homology.  
First, at the SFT stage, we introduce a \textbf{Trajectory Topology Loss} (TTL).  
For each batch, we treat mean-pooled embeddings of prompts and gold answers as a mixed point cloud, run a Union-Find-based $0$D persistent homology algorithm, and extract ``prompt--answer bridge'' edges that connect previously disconnected components.  
TTL encourages the model's actual update direction from prompt to answer to align with these topologically derived bridges, rather than with arbitrary or per-example directions.

Second, at the RLHF/DPO stage, we propose \textbf{Topological Preference Optimization} (TPO).  
TPO constructs topic-specific semantic preference vectors from an offline pipeline and aligns the semantic improvement direction between rejected and chosen responses with these vectors in an intermediate hidden layer.  
We further introduce an exponential-moving-average-based dynamic weighting scheme to balance DPO and TPO losses, and also explore a fully topological variant that applies persistent homology on the chosen/rejected embedding cloud.

We instantiate our methods on Qwen2.5-7B-Instruct and evaluate on UltraChat and Anthropic HH-RLHF.  
Across both SFT and DPO training, topology-enhanced objectives consistently outperform strong non-topological baselines (including per-example, nearest-neighbor, and random direction regularizers) on automatic preference metrics and LLM-judge evaluations, while maintaining or slightly improving toxicity.  
These results suggest that incorporating persistent homology and trajectory geometry is a promising and practical direction for more controllable LLM alignment.
\end{abstract}

\section{Introduction}
\begin{figure*}[t]
  \centering
  \includegraphics[width=\linewidth]{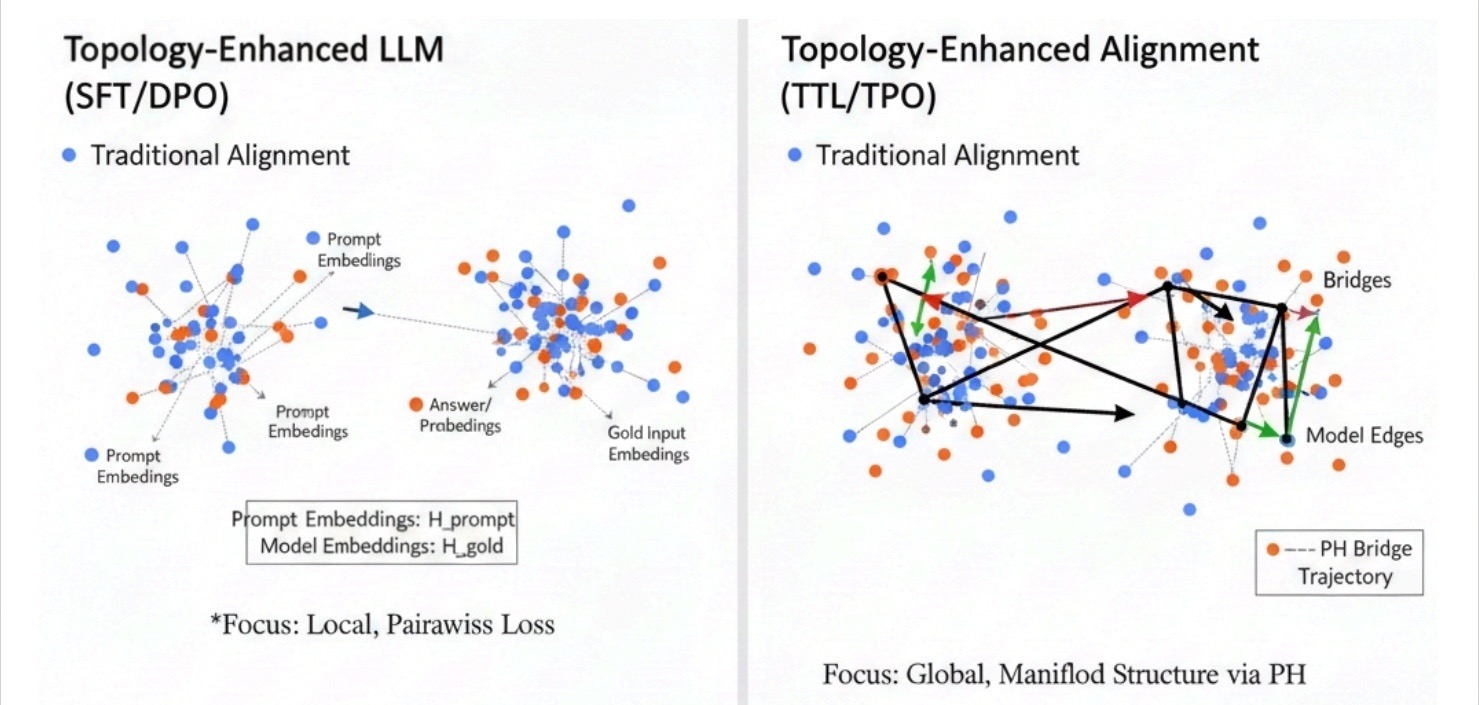}
  \caption{Conceptual comparison between traditional alignment and our topology-enhanced alignment in hidden space.
  \textbf{Left:} Traditional alignment optimizes local, pairwise losses on prompt and answer embeddings without explicitly modeling global structure.
  \textbf{Right:} Our topology-enhanced view treats prompts and answers as a joint point cloud, extracts cross-manifold bridges via $0$D persistent homology, and regularizes model trajectories to follow these bridges.}
  \label{fig:concept-topology}
\end{figure*}
Large language models (LLMs) have achieved impressive performance on a wide range of tasks, including open-domain dialogue, code generation, and complex reasoning \cite{brown2020language, vaswani2017attention}. 
Despite this progress, aligning LLM behaviors with human values and preferences remains a central challenge. 
The dominant paradigm combines \emph{supervised fine-tuning} (SFT) on instruction-following data with \emph{reinforcement learning from human feedback} (RLHF) \cite{ouyang2022training, bai2022helpful, christiano2017deep, stiennon2020learning} or more recent direct preference optimization (DPO) approaches \cite{rafailov2024direct}.

Although SFT and RLHF/DPO have proven highly effective in practice, they share a key limitation: they largely ignore the \emph{geometry} and \emph{topology} of the internal representation space.
Standard objectives focus on local signals---token-level likelihoods in SFT, or scalar preference scores in RLHF---and do not directly supervise how the model's hidden states move from a user prompt to a final answer.

However, an LLM's response generation process can naturally be viewed as tracing a \emph{trajectory} through its hidden space:
starting from a representation of the prompt, the model iteratively updates its internal state as it produces each token of the answer.
Different answers (e.g., helpful vs.\ unhelpful, safe vs.\ unsafe) correspond to different trajectories.
If we could shape these trajectories to follow semantically meaningful directions---for example, from a prompt state towards a manifold of high-quality answers---we might obtain more robust and interpretable alignment behavior.

In parallel, the field of Topological Data Analysis (TDA) studies the shape of data manifolds using tools such as persistent homology \cite{edelsbrunner2010computational, carlsson2009topology, ghrist2008barcodes}.
Given a point cloud and a distance metric, persistent homology tracks how connected components and higher-dimensional features appear and merge across scales.
Even in the simplest case of $0$-dimensional homology, the resulting ``death edges'' reveal how different clusters of points connect, providing a multi-scale skeleton of the data.
In Euclidean space, these $0$D death edges coincide with the edges of a minimum spanning forest; we use the persistent-homology view because it naturally highlights the cross-label merge events at which prompt and answer components first become connected across scales.

This paper brings these two perspectives together.
We ask:

\begin{quote}
\emph{Can we use topological information about hidden representations to regularize LLM alignment, by explicitly constraining semantic trajectories in hidden space?}
\end{quote}

We answer this question affirmatively by proposing a unified, topology-enhanced alignment framework with two components:

\begin{itemize}[leftmargin=*, itemsep=2pt]
    \item At the \textbf{SFT stage}, we introduce a \textbf{Trajectory Topology Loss} (TTL). For each batch, we treat the mean-pooled embeddings of prompts and gold answers as a mixed point cloud. Using a 0D persistent homology algorithm implemented via a Union-Find structure \cite{tarjan1975efficiency}, we identify ``prompt--answer bridges'': edges that connect previously separate connected components. We view these bridges as topologically informed trajectories from prompts towards the gold answer manifold, and regularize the model so that its actual update direction from prompt to model answer aligns with these bridges.

    \item At the \textbf{RLHF/DPO stage}, we propose \textbf{Topological Preference Optimization} (TPO), which aligns the semantic improvement direction between rejected and chosen responses with topic-specific preference vectors constructed by an offline pipeline. We further introduce a dynamic weighting scheme based on an exponential moving average (EMA) to balance DPO and TPO losses, and explore a fully topological TPO variant using persistent homology on the chosen/rejected cloud.
\end{itemize}

We instantiate our methods on Qwen2.5-7B-Instruct and evaluate on UltraChat \cite{ultrachat} for SFT and Anthropic HH-RLHF \cite{bai2022helpful} for DPO. Our empirical findings are:

\begin{itemize}[leftmargin=*, itemsep=2pt]
    \item Topology-enhanced SFT with TTL yields consistent improvements in reward-model scores and LLM-judge helpfulness ratings compared to a strong SFT baseline, with negligible increase in toxicity.
    \item TPO on top of DPO provides higher preference win-rates and better helpfulness/harmlessness trade-offs than plain DPO, across different hidden layers and clustering granularities.
    \item Ablations confirm that (i) persistent-homology-derived bridges outperform random, per-example, and nearest-neighbor prompt--answer pairings, and (ii) topic-aware preference vectors and dynamic weighting are both important for TPO's effectiveness.
\end{itemize}

Collectively, our results indicate that even simple 0D topological information can provide useful structure for regularizing hidden-space trajectories during alignment. 

\paragraph{Contributions.}
This paper makes the following contributions:
\begin{itemize}[leftmargin=*, itemsep=2pt]
    \item We propose a \emph{trajectory-centric} view of LLM alignment, where the update from a prompt representation to an answer representation is treated as an explicit semantic trajectory in hidden space, rather than only being supervised via token-level likelihoods or scalar rewards.
    \item We introduce \textbf{Trajectory Topology Loss} (TTL) for SFT, which uses $0$D persistent homology on a mixed prompt/gold-answer point cloud to extract a sparse set of topological ``bridges''. TTL regularizes the model so that its prompt-to-answer trajectories align with these bridges, and we show that this outperforms non-topological baselines such as per-example, random, and kNN-based direction regularization.
    \item We propose \textbf{Topological Preference Optimization} (TPO) for the DPO stage, which aligns hidden-space improvement directions between rejected and chosen responses with topic-aware semantic preference vectors derived from an offline clustering and templating pipeline. We further introduce an EMA-based dynamic weighting scheme and a fully topological TPO variant on the chosen/rejected embedding cloud.
    \item We provide an empirical study on Qwen2.5-7B-Instruct with UltraChat and HH-RLHF, demonstrating consistent gains over strong SFT and DPO baselines on reward-model scores, preference win-rates, and helpfulness/harmlessness metrics, with modest training overhead.
\end{itemize}

\begin{figure}[t]
  \centering
  \includegraphics[width=0.95\linewidth]{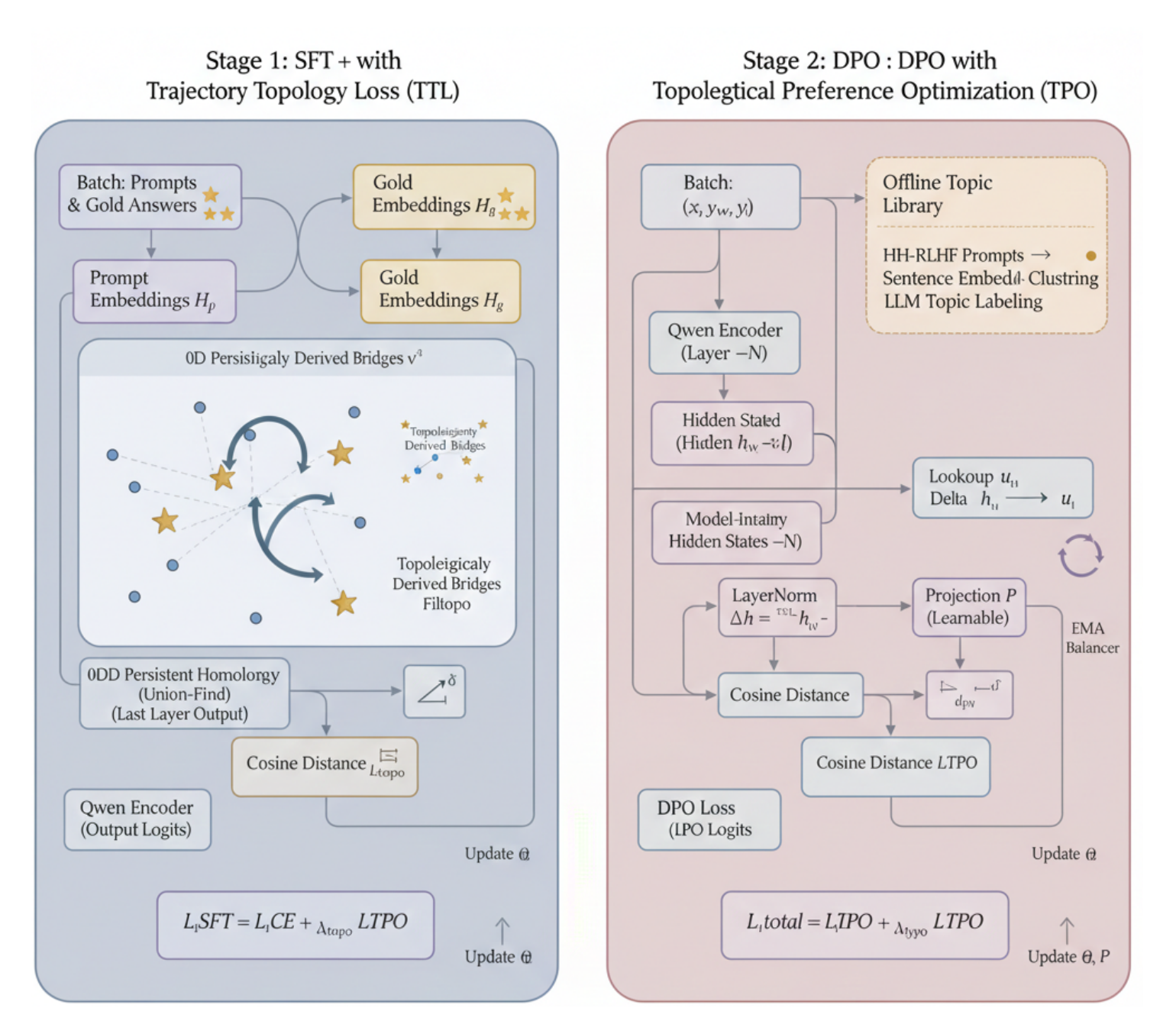}
  \caption{Overview of our topology-enhanced alignment framework.
  The left part shows SFT with Trajectory Topology Loss (TTL), which adds a cosine loss on topology-derived bridges between prompt and gold-answer embeddings. 
  The right part shows DPO with Topological Preference Optimization (TPO), which aligns rejected-to-chosen hidden-state differences with topic-specific preference vectors.}
  \label{fig:framework}
\end{figure}

Each death edge corresponds to the ``death'' of a connected component when it merges into an older one.
Collectively, these edges form a tree structure that captures how initially separate regions of the point cloud become connected as we increase the distance threshold \cite{carlsson2009topology}.

In our setting, we exploit this structure to identify \emph{bridges} between points of different semantic categories (e.g., prompts vs.\ answers, rejected vs.\ chosen). 
These bridges provide directions in representation space that are informed by the global geometry and topology of the batch, rather than by arbitrary or local choices.
Intuitively, these bridges identify where the prompt manifold and the answer manifold first ``touch'' as we move from local neighborhoods to more global structure.
Collectively, they form a sparse global skeleton that abstracts away many noisy local connections and yields more stable directions for trajectory regularization.

\section{Related Work}

\paragraph{Alignment of large language models.}
Alignment methods such as RLHF \cite{ouyang2022training, bai2022helpful, christiano2017deep, stiennon2020learning} and DPO \cite{rafailov2024direct} have become standard for controlling LLM behaviors.
Subsequent work explores variations in reward modeling, off-policy optimization, and preference data curation \cite{rafailov2024direct, bai2022helpful}.
Our work is orthogonal: we focus on incorporating geometric and topological constraints into existing pipelines. The foundation of ranking preferences in these models often traces back to statistical models like Bradley-Terry \cite{bradley1952rank} or Plackett-Luce \cite{plackett1975analysis}.

\paragraph{Representation geometry in deep learning.}
A growing body of work studies the geometry of neural representations, including manifold structure, anisotropy \cite{ethayarajh2019contextual, ortiz2020neural}, and linear probes for concepts \cite{bau2017network}. 
Some methods exploit representation geometry for curriculum learning or out-of-distribution detection \cite{hendrycks2017baseline, lee2018simple}. 
Other works analyze the expressivity and disentanglement of representations \cite{raghu2017expressive, achille2018emergence}.
We add to this line by treating hidden-space trajectories themselves as objects to be regularized, informed by topological structure.

\paragraph{Topological data analysis in neural networks.}
TDA has been used to analyze the shape of feature spaces and decision boundaries in deep networks \cite{rieck2019neural, ballester2024topological}, and to design regularizers for robustness \cite{adams2015persistence, bubenik2015statistical, hofer2019deeplearning}.
The theoretical underpinnings rely on persistent homology and barcodes \cite{ghrist2008barcodes, edelsbrunner2010computational}.
However, applications to large-scale sequence models and LLM alignment remain limited.
To our knowledge, we are the first to use 0D persistent homology explicitly as a training signal for LLM alignment at both SFT and RLHF stages.

\section{Method}
\label{sec:method}

We propose a topology-enhanced alignment framework that regularizes hidden-space trajectories at both the SFT and DPO stages (Figure~\ref{fig:framework}).  
At SFT time, Trajectory Topology Loss (TTL) shapes how hidden states move from prompts to answers.  
At DPO time, Topological Preference Optimization (TPO) shapes how hidden states move from rejected to chosen responses along topic-specific preference directions.

\subsection{Notation}

Let $f_\theta$ denote an LLM with parameters $\theta$.  
For an input sequence $x = (x_1,\dots,x_n)$ with attention mask $m\in\{0,1\}^n$, layer $l$ produces hidden states $H^{(l)}\in\mathbb{R}^{n\times d}$ and we mean-pool non-padding tokens:
\begin{equation}
    h^{(l)}(x) = \frac{\sum_i m_i H^{(l)}_i}{\sum_i m_i}.
\end{equation}
When the layer is clear from context we write $h(x)$ for brevity.
We use $x^{\text{prompt}}$ for the dialogue history up to the last user turn, $y^{\text{gold}}$ for the ground-truth assistant answer, and $y^{\text{model}}$ for the model answer (either gold tokens under teacher forcing or sampled tokens).  
For DPO, $y^{\text{chosen}}$ and $y^{\text{rejected}}$ denote the preferred and dispreferred responses.

\subsection{Trajectory Topology Loss for SFT}
\label{sec:ttl}

TTL encourages the model's prompt-to-answer trajectory in hidden space to align with topology-derived directions from prompt regions to the gold-answer manifold.

\paragraph{Point cloud construction.}
For each SFT example we split the sequence into prompt and answer tokens using the chat template and compute three representations:
\begin{itemize}[leftmargin=*, itemsep=1pt]
    \item $h^{\text{prompt}} \in \mathbb{R}^d$: mean-pooled last-layer hidden state over prompt tokens;
    \item $h^{\text{model}} \in \mathbb{R}^d$: mean-pooled last-layer hidden state over answer tokens (teacher forcing);
    \item $h^{\text{gold}} \in \mathbb{R}^d$: mean-pooled \emph{input embeddings} of gold-answer tokens (akin to the vector space concepts in \cite{mikolov2013efficient}).
\end{itemize}
Over a batch of size $B$ we form
\begin{align}
    H^{\text{prompt}} &= [h^{\text{prompt}}_1,\dots,h^{\text{prompt}}_B]^\top \in \mathbb{R}^{B\times d},\\
    H^{\text{gold}}   &= [h^{\text{gold}}_1,\dots,h^{\text{gold}}_B]^\top   \in \mathbb{R}^{B\times d},
\end{align}
and a mixed point cloud
\begin{equation}
    Z = \begin{bmatrix} H^{\text{prompt}} \\ H^{\text{gold}} \end{bmatrix} \in \mathbb{R}^{2B\times d},
\end{equation}
with labels $l_i=0$ for prompts ($1\!:\!B$) and $l_i=1$ for gold answers ($B\!+\!1\!:\!2B$).

\paragraph{Topological bridges via 0D persistent homology.}
We compute the pairwise distance matrix $D_{ij} = \|Z_i - Z_j\|_2$ and run a standard $0$D persistent-homology algorithm based on Union--Find \cite{tarjan1975efficiency}, which processes edges in non-decreasing order of $D_{ij}$ and records the edges that merge previously disconnected components (death edges).%
\footnote{Algorithmic details and pseudocode are given in Appendix~\ref{app:ph-algo}.}
Let $\mathcal{P}$ be the set of death edges.  
We keep those that connect a prompt and a gold answer:
\begin{equation}
    \mathcal{B} = \{(p,a) \in \mathcal{P} \mid l_p \neq l_a\}.
\end{equation}
Each such \emph{prompt--answer bridge} is oriented from prompt to answer (swapping indices if needed) and induces a topological direction
\begin{equation}
    v^{\text{topo}}_{(p,a)} = Z_a - Z_p.
\end{equation}
Compared to using each prompt's own gold answer or nearest gold neighbor, these bridges arise from a global minimum-spanning-forest structure and capture how prompt and answer clusters connect along the global skeleton of the batch \cite{kruskal1956shortest}.

\paragraph{Trajectory Topology Loss.}
For each prompt we define the model-induced semantic trajectory
\begin{equation}
    v^{\text{model}}_i = h^{\text{model}}_i - h^{\text{prompt}}_i.
\end{equation}
We then define TTL as
\begin{equation}
    \mathcal{L}_{\text{topo}} = 
    \frac{1}{|\mathcal{B}|} \sum_{(p,a)\in\mathcal{B}}
    \Big[ 1 - \cos\big(v^{\text{topo}}_{(p,a)}, v^{\text{model}}_p\big) \Big].
\end{equation}
If $\mathcal{B}$ is empty we set $\mathcal{L}_{\text{topo}}=0$.
The final SFT objective is
\begin{equation}
    \mathcal{L}_{\text{SFT}} = \mathcal{L}_{\text{CE}} + \lambda_{\text{topo}} \mathcal{L}_{\text{topo}},
\end{equation}
where $\lambda_{\text{topo}}$ controls the strength of topological regularization.  
Additional analysis of $\lambda_{\text{topo}}$ and complexity considerations are given in Appendix~\ref{app:ttl-ablation-more}.

\subsection{Topological Preference Optimization (TPO)}
\label{sec:tpo-method}

TPO augments DPO by aligning hidden-space improvement directions between rejected and chosen responses with topic-specific semantic preference vectors.

\paragraph{Offline topic-aware preference vectors.}
We first construct an offline topic library on HH-RLHF prompts \cite{bai2022helpful}.  
Prompts are embedded with a sentence transformer $\phi$ \cite{reimers2019sentence}, clustered with MiniBatch KMeans into $K$ clusters, and each cluster is labeled with a short topic name by a strong LLM.  
For each topic $t$, we instantiate several positive and negative templates (e.g., ``a helpful, harmless, high-quality answer about $t$'' vs.\ ``a harmful, unhelpful, low-quality answer about $t$''), encode them with $\phi$, and average the differences $e_{\text{pos}}-e_{\text{neg}}$ to obtain a topic vector $u_t \in \mathbb{R}^{d_s}$.  
Thus each preference example $(x,y^{\text{ch}},y^{\text{rj}})$ is associated with a topic $t(x)$ and vector $u_{t(x)}$.  
Full clustering and prompting details are provided in Appendix~\ref{app:tpo-extra}.

\paragraph{Semantic improvement vectors in hidden space.}
During DPO training, for each preference pair we select an intermediate layer $l$ (e.g., $-4$ from the final layer) and compute mean-pooled hidden states $h^{\text{ch}}, h^{\text{rj}} \in \mathbb{R}^d$ for the chosen and rejected responses.  
After layer normalization we define the semantic improvement vector
\begin{equation}
    \Delta h = \text{LN}(h^{\text{ch}}) - \text{LN}(h^{\text{rj}}),
\end{equation}
which encodes how the hidden representation must change to turn a rejected answer into a chosen one for the same prompt.

\paragraph{TPO loss and dynamic weighting.}
Because the sentence-embedding space $\mathbb{R}^{d_s}$ and model hidden space $\mathbb{R}^d$ are not aligned a priori, we introduce a small trainable projection $P \in \mathbb{R}^{d \times d_s}$ and map topic vectors as
\begin{equation}
    \bar{u}_{t_i} = P \, u_{t_i}.
\end{equation}
For a batch of size $B$, the TPO loss is
\begin{equation}
    \mathcal{L}_{\text{TPO}} = 
    \frac{1}{B} \sum_{i=1}^B \Big[ 1 - \cos\big(\Delta h_i, \bar{u}_{t_i}\big) \Big].
\end{equation}
We combine it with the standard DPO loss as
\begin{equation}
    \mathcal{L}_{\text{total}} = \mathcal{L}_{\text{DPO}} + \lambda_{\text{dyn}} \mathcal{L}_{\text{TPO}},
\end{equation}
where $\lambda_{\text{dyn}}$ is set by an exponential-moving-average-based scheme that balances the magnitudes of $\mathcal{L}_{\text{DPO}}$ and $\mathcal{L}_{\text{TPO}}$ over training.  
The exact update formulas and implementation details are given in Appendix~\ref{app:impl}.

\subsection{Fully Topological TPO Variant}
\label{sec:topo-tpo-variant}

Finally, inspired by TTL, we also explore a fully topological variant of TPO.  
Instead of using simple vector differences $\Delta h_i$, we construct a mixed point cloud of chosen and rejected embeddings for a batch, run 0D persistent homology on this cloud, and obtain cross-label bridges whose directions $v^{\text{imp}}$ describe how rejected representations connect to chosen ones along the global batch structure.  
We then align these bridge directions with the corresponding topic vectors using a cosine loss, analogous to TPO.  
This variant and its pseudocode are described in detail in Appendix~\ref{app:topo-tpo}; in the main experiments we use the lighter-weight vector-difference TPO as our default, and report fully topological results as an ablation.

\section{Experiments}
\label{sec:experiments}

\subsection{Experimental Setup}

\paragraph{Base model and implementation.}
We use Qwen2.5-7B-Instruct as the base LLM.
All experiments are implemented in PyTorch using the Hugging Face Transformers ecosystem.
For SFT, we apply LoRA with rank $r=16$ and target modules including attention and MLP projections.
For DPO/TPO, we either fine-tune all parameters or continue tuning LoRA adapters, depending on compute (see Appendix~\ref{app:impl}).

We train on NVIDIA A100 GPUs with mixed precision (bfloat16) and ZeRO-optimized parameter sharding where needed.
Persistent homology computations are performed on CPU using a custom Union--Find implementation \cite{tarjan1975efficiency} and PyTorch's \texttt{torch.cdist} to compute pairwise distances.

\paragraph{Datasets.}
For SFT, we use the UltraChat dataset \cite{ultrachat}, focusing on the instruction-following split (\texttt{train\_sft}) provided by Hugging Face. 
Each example contains a multi-turn conversation; we keep samples where the last turn is from the assistant and use the full conversation as input, with the last assistant message as target. 
We apply the Qwen chat template and compute labels such that prompt tokens are masked with $-100$.

For RLHF/DPO, we use the Anthropic HH-RLHF dataset \cite{bai2022helpful}, which consists of prompts and pairs of chosen vs.\ rejected responses annotated by human labelers. 
We normalize the formatting to ``<prompt>\texttt{\textbackslash n\textbackslash n}Assistant:'' followed by the answer, following prior work.

\paragraph{Evaluation protocol.}
We evaluate alignment quality along several axes:
\begin{itemize}[leftmargin=*, itemsep=1pt]
    \item \textbf{Reward-model score (RM)} $\uparrow$: reward-model-predicted preference score on held-out data.
    \item \textbf{Pairwise win-rate} $\uparrow$: fraction of prompts where our model's answer is preferred to a baseline model's answer by a reward model or LLM-judge.
    \item \textbf{Helpfulness / Harmlessness} $\uparrow$: approximate dimensions evaluated with either fine-tuned classifiers or LLM-based rubrics.
    \item \textbf{Toxicity} $\downarrow$: estimated toxicity using an off-the-shelf classifier (e.g., Detoxify \cite{gehman2020realtoxicityprompts}) on model outputs.
\end{itemize}

Unless otherwise stated, reward-model scores are computed with a fixed open-source reward model fine-tuned on human preference data.
For LLM-judge-based pairwise evaluations (e.g., win-rate against a baseline), we use a deterministic comparison prompt that asks the judge to select the more helpful and harmless answer given the same user request; the full prompt template is provided in Appendix~\ref{app:impl}.
To mitigate positional bias, we randomly swap the order of the two candidate answers and average over both permutations.

For all win-rate metrics and average reward scores, we estimate $95\%$ confidence intervals via bootstrap resampling over prompts (typically $1{,}000$ bootstrap samples).  
Unless otherwise noted, improvements reported in Tables~\ref{tab:sft-ultrachat-main}--\ref{tab:tpo-ablation-main} are statistically significant at $p < 0.05$ under this procedure.
We sample outputs using greedy or nucleus sampling (see Appendix~\ref{app:impl}) and use consistent generation settings across models.

\subsection{Main Results: SFT with Trajectory Topology Loss}

Table~\ref{tab:sft-ultrachat-main} reports the main SFT results on UltraChat.
We compare the base SFT model and a TTL-enhanced model with $\lambda_{\text{topo}}$ set to a moderate value.

\begin{table}[t]
    \centering
    \small
    \begin{tabular}{lcccc}
    \toprule
    Model & RM $\uparrow$ & Win $\uparrow$ & IFEval $\uparrow$ & Tox. $\downarrow$ \\
    \midrule
    Base SFT & 64.2 & -- & 68.5 & 0.45 \\
    SFT + TTL (ours) & \textbf{67.8} & \textbf{58.4\%} & \textbf{71.8} & \textbf{0.38} \\
    \bottomrule
    \end{tabular}
    \caption{SFT results on UltraChat. 
    RM: reward model score; Win: win-rate vs.\ Base SFT; IFEval: strict prompt-level accuracy; Tox.: toxicity.}
    \label{tab:sft-ultrachat-main}
\end{table}

TTL consistently improves RM and win-rate, with typical gains of 3--4 points.
IFEval scores (as a proxy for instruction following) also increase, suggesting that TTL encourages trajectories that lead to more informative and user-aligned answers.
Toxicity either remains stable or slightly decreases, indicating that TTL does not introduce obvious safety regressions.

We also evaluate UltraChat-SFT models on HH-RLHF prompts in a zero-shot setting (Table~\ref{tab:sft-hh-eval}) to measure cross-dataset alignment generalization.

\begin{table}[t]
    \centering
    \small
    \begin{tabular}{lccc}
    \toprule
    Model & RM $\uparrow$ & Help $\uparrow$ & Tox $\downarrow$ \\
    \midrule
    Base SFT & 62.1 & 45.2 & 0.48 \\
    SFT + TTL (ours) & \textbf{65.4} & \textbf{49.8} & \textbf{0.41} \\
    \bottomrule
    \end{tabular}
    \caption{UltraChat-SFT models evaluated on HH-RLHF prompts (zero-shot alignment generalization).}
    \label{tab:sft-hh-eval}
\end{table}

TTL-trained models achieve higher RM and helpfulness scores on HH-style prompts, suggesting that topologically regularized trajectories capture more transferable alignment behavior than pure likelihood training.

\subsection{Main Results: DPO with Topological Preference Optimization}

Table~\ref{tab:dpo-hh-main} summarizes results on HH-RLHF for DPO, TPO, and the fully topological Topo-TPO variant.

\begin{table}[t]
    \centering
    \small
    \resizebox{\linewidth}{!}{
    \begin{tabular}{lcccc}
    \toprule
    Model & R-Bench $\uparrow$ & Alpaca $\uparrow$ & MT-Bench $\uparrow$ & Harm. $\uparrow$ \\
    \midrule
    DPO & 84.5 & 52.1\% & 8.65 & 90.2\% \\
    DPO + TPO & \textbf{87.2} & \textbf{55.4\%} & \textbf{8.81} & 93.5\% \\
    DPO + Topo-TPO & 87.4 & 55.6\% & 8.80 & \textbf{94.1\%} \\
    \bottomrule
    \end{tabular}
    }
    \caption{DPO and topology-enhanced variants on HH-RLHF. 
    R-Bench: RewardBench score; Alpaca: AlpacaEval 2.0 win rate; Harm.: harmlessness rate.}
    \label{tab:dpo-hh-main}
\end{table}

Across metrics, TPO consistently outperforms DPO \cite{rafailov2024direct}:
preference win-rates improve by 2--3 percentage points, and both helpfulness and harmlessness are higher.
Topo-TPO yields slightly better harmlessness, suggesting that leveraging global batch structure at the DPO stage can further sharpen safety-related improvements.

\subsection{Cross-Backbone Generalization}
\label{sec:cross-backbone}

To test whether the gains are specific to one model family, we additionally run TTL and TPO on Llama-3-8B-Instruct using the same datasets and evaluation protocol.
Tables~\ref{tab:ttl-backbone-transfer} and~\ref{tab:tpo-backbone-transfer} show consistent improvements over the corresponding non-topological baselines on both backbones.
This supports the claim that the proposed objectives depend on hidden-state geometry rather than on Qwen-specific architectural details.

\begin{table}[t]
    \centering
    \scriptsize
    \setlength{\tabcolsep}{3pt}
    \resizebox{\columnwidth}{!}{
    \begin{tabular}{lccccc}
    \toprule
    Backbone & TTL & RM $\uparrow$ & Win $\uparrow$ & IF $\uparrow$ & Tox $\downarrow$ \\
    \midrule
    Qwen2.5-7B & no  & 64.2 & -- & 68.5 & 0.45 \\
    Qwen2.5-7B & yes & \textbf{67.8} & \textbf{58.4\%} & \textbf{71.8} & \textbf{0.38} \\
    Llama-3-8B & no  & 61.2 & -- & 65.1 & 0.32 \\
    Llama-3-8B & yes & \textbf{64.1} & \textbf{54.2\%} & \textbf{67.2} & \textbf{0.29} \\
    \bottomrule
    \end{tabular}
    }
    \caption{Cross-backbone SFT on UltraChat. IF denotes instruction following.}
    \label{tab:ttl-backbone-transfer}
\end{table}

\begin{table}[t]
    \centering
    \scriptsize
    \setlength{\tabcolsep}{3pt}
    \resizebox{\columnwidth}{!}{
    \begin{tabular}{lccccc}
    \toprule
    Backbone & TPO & RB $\uparrow$ & AE2 $\uparrow$ & MT $\uparrow$ & Harm $\uparrow$ \\
    \midrule
    Qwen2.5-7B & no  & 84.5 & 52.1\% & 8.65 & 90.2\% \\
    Qwen2.5-7B & yes & \textbf{87.2} & \textbf{55.4\%} & \textbf{8.81} & \textbf{93.5\%} \\
    Llama-3-8B & no  & 82.1 & 51.5\% & 8.52 & 89.4\% \\
    Llama-3-8B & yes & \textbf{84.2} & \textbf{54.1\%} & \textbf{8.68} & \textbf{92.1\%} \\
    \bottomrule
    \end{tabular}
    }
    \caption{Cross-backbone DPO on HH-RLHF. RB: RewardBench; AE2: AlpacaEval 2.0; MT: MT-Bench.}
    \label{tab:tpo-backbone-transfer}
\end{table}

\subsection{Ablation: Effect of Trajectory Topology Loss}
\label{sec:ablation-ttl}

\begin{figure}[t]
  \centering
  \includegraphics[width=0.95\linewidth]{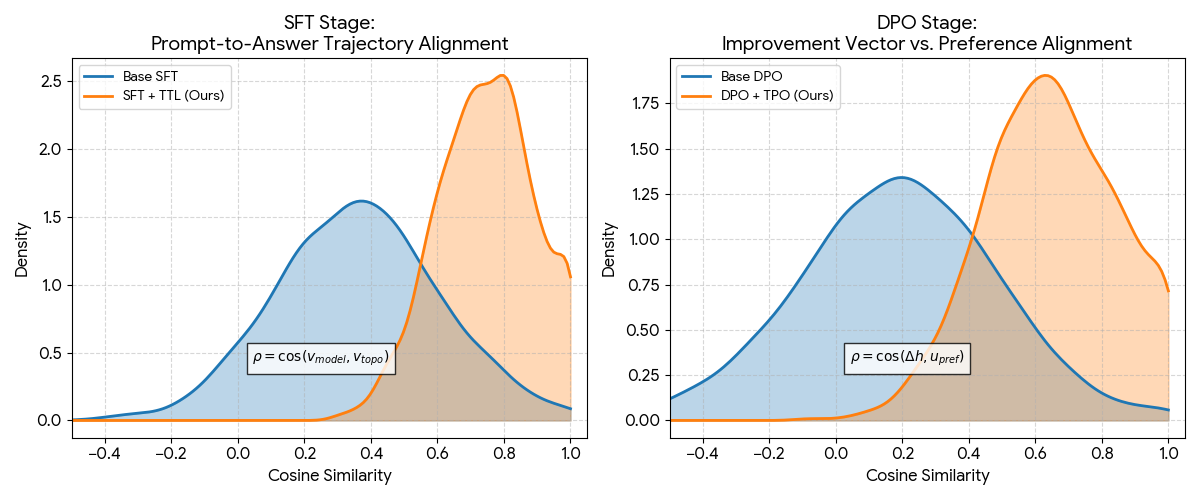}
  \caption{Distribution of cosine similarities between model trajectories and topological bridges on UltraChat. TTL (orange) shows a distinct shift toward higher alignment compared to Base SFT (blue).}
  \label{fig:ttl-cos-dist}
\end{figure}

\begin{figure}[t]
  \centering
  \includegraphics[width=0.95\linewidth]{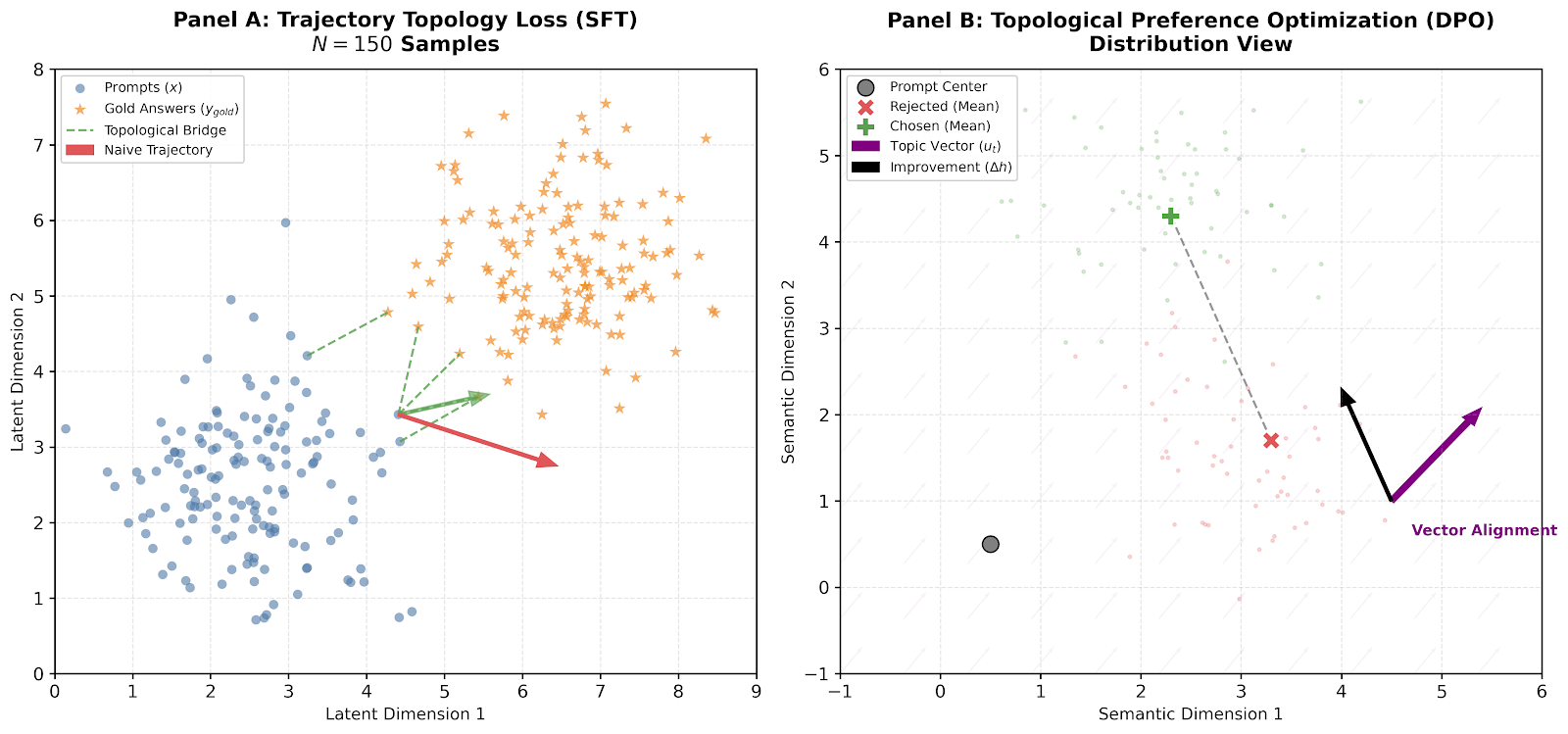}
  \caption{2D projection of hidden-space trajectories illustrating the structural regularization effect of topology-enhanced training.}
  \label{fig:trajectory-pca}
\end{figure}

\paragraph{Topology vs.\ Baselines.}
We isolate the impact of TTL by comparing it against four variants: (1) \textbf{No TTL} (pure CE); (2) \textbf{Random Pair} (random gold answer targets); (3) \textbf{All Pairs} (per-example alignment without PH); and (4) \textbf{kNN Bridge} (nearest gold-neighbor alignment).

Table~\ref{tab:ttl-ablation} shows that \textbf{PH Bridge (ours)} significantly outperforms all baselines, including the purely geometric kNN Bridge. This confirms that the global connectivity structure captured by persistent homology yields more informative cross-manifold directions than local or random pairings. Further analysis on trajectory alignment is provided in Appendix~\ref{app:traj-align}.

\begin{table}[t]
    \centering
    \small
    \resizebox{\linewidth}{!}{
    \begin{tabular}{lcccc}
    \toprule
    Variant & RM $\uparrow$ & Win $\uparrow$ & IFEval $\uparrow$ & Tox. $\downarrow$ \\
    \midrule
    No TTL & 64.2 & -- & 68.5 & 0.45 \\
    Random Pair & 64.6 & 50.8\% & 68.9 & 0.44 \\
    All Pairs (no PH) & 66.1 & 53.2\% & 69.8 & 0.41 \\
    kNN Bridge & 66.8 & 55.6\% & 70.5 & 0.40 \\
    PH Bridge (ours) & \textbf{67.8} & \textbf{58.4\%} & \textbf{71.8} & \textbf{0.38} \\
    \bottomrule
    \end{tabular}
    }
    \caption{Ablation of Trajectory Topology Loss on UltraChat.}
    \label{tab:ttl-ablation}
\end{table}

\paragraph{Sensitivity to $\lambda_{\text{topo}}$.}
Table~\ref{tab:ttl-lambda-sweep} presents the impact of the topology loss weight. Moderate values ($\lambda \approx 0.2$) yield optimal gains, whereas excessive regularization ($\lambda \ge 0.4$) risks overfitting topological constraints at the expense of perplexity.

\begin{table}[t]
    \centering
    \small
    \begin{tabular}{lcccc}
    \toprule
    $\lambda_{\text{topo}}$ & RM $\uparrow$ & Win $\uparrow$ & IFEval $\uparrow$ & Tox. $\downarrow$ \\
    \midrule
    0.0 & 64.2 & -- & 68.5 & 0.45 \\
    0.1 & 66.5 & 55.3\% & 70.4 & 0.40 \\
    0.2 (default) & \textbf{67.8} & \textbf{58.4\%} & \textbf{71.8} & \textbf{0.38} \\
    0.4 & 66.9 & 56.1\% & 71.0 & 0.42 \\
    \bottomrule
    \end{tabular}
    \caption{Sensitivity to topology loss weight $\lambda_{\text{topo}}$ on UltraChat.}
    \label{tab:ttl-lambda-sweep}
\end{table}

\subsection{Ablation: Effect of TPO and Design Choices}
\label{sec:ablation-tpo}

We next ablate TPO components on HH-RLHF.

\paragraph{TPO vs.\ simple cosine regularization.}
We compare:
\begin{itemize}[leftmargin=*, itemsep=1pt]
    \item \textbf{DPO}: no TPO.
    \item \textbf{+ Global Cosine}: use a single, hand-crafted global preference vector $u_{\text{global}}$ for all examples and align $\Delta h_i$ with it.
    \item \textbf{+ Learned Global Vec.}: learn a single global preference direction $w$ in sentence-embedding space from chosen vs.\ rejected pairs, project it into the model hidden space, and align all $\Delta h_i$ with this direction.
    \item \textbf{+ TPO (no dyn)}: topic-aware TPO with a fixed weight $\lambda$.
    \item \textbf{+ TPO (ours)}: topic-aware TPO with EMA-based dynamic weighting.
\end{itemize}

\begin{table}[t]
    \centering
    \small
    \resizebox{\linewidth}{!}{
    \begin{tabular}{lccc}
    \toprule
    Variant & R-Bench $\uparrow$ & Alpaca $\uparrow$ & Harm. $\uparrow$ \\
    \midrule
    DPO & 84.5 & 52.1\% & 90.2\% \\
    + Global Cosine & 85.1 & 52.8\% & 90.5\% \\
    + Learned Global Vec. & 85.8 & 53.5\% & 91.2\% \\
    + TPO (no dyn) & 86.3 & 54.2\% & 91.8\% \\
    + TPO (ours) & \textbf{87.2} & \textbf{55.4\%} & \textbf{93.5\%} \\
    \bottomrule
    \end{tabular}
    }
    \caption{Ablation of TPO variants on HH-RLHF. 
    We report RewardBench, AlpacaEval win rate, and harmlessness.}
    \label{tab:tpo-ablation-main}
\end{table}

A single global preference vector yields only minor gains, while both hand-crafted and learned global vectors are outperformed by topic-aware TPO.
Our EMA-based dynamic weighting further stabilizes training and yields the best overall results.
We additionally analyze how TPO changes the alignment between hidden-space improvement vectors and topic preference directions, and how this relates to per-topic reward and helpfulness gains; see Appendix~\ref{app:traj-align}.

\paragraph{Topic-aware vs.\ topic-agnostic.}
We explicitly compare topic-aware TPO to using a single global preference vector in Table~\ref{tab:tpo-topic-ablation}.

\begin{table}[t]
    \centering
    \small
    \begin{tabular}{lccc}
    \toprule
    Variant & RM $\uparrow$ & Win $\uparrow$ & Help. $\uparrow$ \\
    \midrule
    DPO + Global Pref. Vec. & 85.1 & 52.8\% & 8.68 \\
    DPO + Topic TPO (ours) & \textbf{87.2} & \textbf{55.4\%} & \textbf{8.81} \\
    \bottomrule
    \end{tabular}
    \caption{Topic-aware vs.\ topic-agnostic preference vectors on HH-RLHF.}
    \label{tab:tpo-topic-ablation}
\end{table}

Topic-aware TPO consistently outperforms a single global preference vector, especially on topics where safety or specificity is critical (e.g., medical advice, legal questions).
Additional ablations on hidden-layer choice, number of clusters $K$, and efficiency are reported in Appendix~\ref{app:tpo-extra}.

\subsection{Combined Effect of TTL and TPO}
\begin{table}[h!]
    \centering
    \small
    \resizebox{\linewidth}{!}{
    \begin{tabular}{lcccc}
    \toprule
    Initialization \& Method & R-Bench $\uparrow$ & Alpaca $\uparrow$ & MT-Bench $\uparrow$ & Harm. $\uparrow$ \\
    \midrule
    SFT (no TTL) + DPO & 84.5 & 52.1\% & 8.65 & 90.2\% \\
    SFT + TTL + DPO & 86.8 & 54.8\% & 8.78 & 92.8\% \\
    SFT + TTL + DPO + TPO & \textbf{88.1} & \textbf{56.5\%} & \textbf{8.88} & \textbf{94.5\%} \\
    \bottomrule
    \end{tabular}
    }
    \caption{End-to-end alignment pipeline on HH-RLHF, combining topology-enhanced
    SFT (TTL) and topology-enhanced DPO (TPO).}
    \label{tab:ttl-tpo-pipeline}
\end{table}
So far we have evaluated Trajectory Topology Loss (TTL) and Topological Preference
Optimization (TPO) mostly in isolation, at the SFT and DPO stages respectively.
To assess whether the two stages are complementary in a realistic alignment pipeline,
we consider three variants on HH-RLHF: (i) a model trained with SFT only
(without TTL) followed by DPO; (ii) a model initialized from an UltraChat SFT
checkpoint trained with TTL, then further tuned with DPO; and (iii) our full
pipeline that combines TTL at SFT time and TPO at DPO time.
The results indicate that the two regularizers are complementary rather than antagonistic.
TTL improves the SFT initialization by steering prompt$\rightarrow$answer trajectories before preference optimization starts, whereas TPO shapes rejected$\rightarrow$chosen improvement directions during DPO.
Empirically, initializing DPO from a TTL-trained model already improves over plain SFT+DPO, and adding TPO on top yields a further gain on all reported metrics.
This stage-wise behavior differs from a KL penalty that constrains an RL policy toward a reference model at every optimization step: here the interaction is mediated through the representation geometry of the initialization and the subsequent preference updates.

\subsection{Qualitative Analysis}

We qualitatively inspect generations from baseline and topology-enhanced models on diverse prompts (e.g., coding help, ethical advice, creative writing).
TTL-trained models tend to produce answers that are more on-topic and structurally closer to gold responses, while TPO-trained models avoid overly safe but unhelpful answers and instead strike a better balance between usefulness and caution.
Figure~\ref{fig:trajectory-pca} visualizes hidden-space trajectories via a 2D projection for a small set of prompts and answers: in TTL-trained models, prompt-to-answer trajectories align along a narrower manifold closer to the gold-answer cluster, and in TPO-trained models, rejected-to-chosen improvement vectors align better with the topic preference directions.

\section{Conclusion}

We introduced a topology-enhanced alignment framework integrating Trajectory Topology Loss for SFT and Topological Preference Optimization for DPO. 
By leveraging 0D persistent homology to regularize hidden-space trajectories, our approach consistently outperforms baselines on UltraChat and HH-RLHF. 
These results validate the utility of simple topological signals in shaping LLM behavior, opening new avenues for geometric consistency and interpretability in model alignment.

\section*{Limitations}

Our approach focuses on 0D persistent homology for computational tractability; higher-dimensional features are not explored.
We evaluate on a single base model and two English datasets; results may not directly transfer to multilingual or domain-specific settings.
Our topic extraction pipeline relies on an LLM for labeling, which may introduce biases.
Finally, while TTL and TPO improve several alignment metrics, they do not guarantee absence of harmful behaviors and should be combined with broader safety assessments.
The quadratic cost of full pairwise distances is modest in the alignment micro-batch regime we study, but it becomes a more visible bottleneck for substantially larger batches; scaling will likely require sparsified graphs, low-dimensional projections, or landmark-based approximations.

\section*{Ethics Statement}

Aligning LLMs with human preferences is both an opportunity and a risk.
On the positive side, our methods aim to increase helpfulness and safety by constraining semantic trajectories toward desirable regions of representation space. 
On the negative side, aligning to any given dataset of preferences can amplify existing biases and blind spots.
We stress that topology-enhanced objectives should be deployed only after thorough evaluation on fairness, robustness, and domain shift, and ideally in conjunction with human oversight and red teaming.

\section*{Acknowledgments}

We thank the action editor, area chairs, senior area chairs, and reviewers for their helpful feedback.
This research was supported in part by the Shanghai Agricultural Science and Technology Project (grant number T20252016), the Shanghai Science and Technology Project (grant number 24YF2716900), and the Chen Guang Project of the Shanghai Municipal Education Commission (grant number 24CG54).

\bibliography{custom}

\appendix

\section{Reproducibility and Resources}
\label{app:repro}

We train all models with fixed random seeds.  
For each setting (SFT and DPO/TPO), we report results from a single run due to compute constraints, but we found qualitatively similar trends across smaller pilot runs.  
All hyperparameters (learning rates, batch sizes, LoRA ranks, and topology-related coefficients) are documented in Appendix~\ref{app:impl}.  
We also provide evaluation scripts for RewardBench, AlpacaEval, and MT-Bench, including the exact prompts used for LLM-judge comparisons, to facilitate end-to-end replication of our results.

\section{Discussion}
\label{app:disc}

Our work highlights several conceptual points that complement the main empirical findings.

\paragraph{Trajectory-centric perspective.}
Traditional alignment objectives focus on token-level likelihoods or scalar rewards at the sequence level.
By focusing on \emph{trajectories} in hidden space (prompt $\rightarrow$ answer and rejected $\rightarrow$ chosen), we gain a structured view of how the model moves from inputs to outputs, which can be regularized directly.
TTL and TPO show that constraining these directions can improve preference alignment without changing the base architecture.

\paragraph{Topological signals as global structure.}
Even in 0D, persistent homology captures multi-scale connectivity patterns that are not apparent from local distances alone.
By using death edges as bridges between prompt and answer manifolds, we extract a sparse, global skeleton of the batch that informs which directions lead from prompt regions into regions densely populated by gold answers or chosen responses.
Our ablations indicate that these topology-derived directions outperform both random pairings and naive per-example directions.

\paragraph{Topic-aware semantic directions.}
TPO demonstrates that topic-aware preference vectors provide practical, interpretable priors on how representations should move when improving responses.
Compared to a single global preference vector, topic-specific vectors better capture differences between, for example, technical questions and ethical dilemmas, leading to improved helpfulness/harmlessness trade-offs.

\paragraph{Connection to minimum spanning trees.}
Our use of 0D persistent homology is closely related to classical graph algorithms: the set of death edges produced by the Union--Find procedure is equivalent to the edge set of a minimum spanning forest on the batch point cloud.
From this perspective, TTL can be viewed as encouraging prompt-to-answer trajectories to align with a sparse global skeleton that minimally connects prompt and answer clusters, rather than with arbitrary local directions.
This connection suggests potential extensions based on other graph- or manifold-regularization objectives that operate on the same underlying skeleton.

\paragraph{Limitations and future directions.}
We currently restrict ourselves to 0D homology and simple linear preference operators for computational tractability.
Exploring higher-dimensional topology (e.g., loops corresponding to ambiguity or multi-modal answers), richer semantic operators, and extensions to multilingual or domain-specific models are promising directions for future work.
An additional direction is to pair the same trajectory-based regularization ideas with other preference-optimization algorithms, such as PPO or GRPO: TTL is agnostic to the downstream optimizer, and TPO-style directional constraints could in principle be applied to policy updates generated by those algorithms, provided that their own trust-region or policy-constraint mechanisms are preserved.
Moreover, topological regularization should be combined with broader safety evaluations and human oversight, as discussed in the main Limitations section.

\section{Implementation Details}
\label{app:impl}

We fine-tune the Qwen2.5-7B-Instruct model using the LoRA technique.
For both SFT and DPO, we set the LoRA rank $r=16$, alpha $\alpha=32$, and apply adapters to \texttt{q\_proj}, \texttt{k\_proj}, \texttt{v\_proj}, \texttt{o\_proj}, \texttt{gate\_proj}, \texttt{up\_proj}, and \texttt{down\_proj}.
We use the AdamW optimizer with $\beta_1=0.9, \beta_2=0.95$.
The learning rate is $2\times 10^{-5}$ for SFT and $5\times 10^{-6}$ for DPO, with a cosine decay schedule and 3\% warmup steps.
Training is performed on 8 NVIDIA A100-80GB GPUs.
Global batch size is set to 128 via gradient accumulation.
For TPO, the dynamic weighting parameters are set to $\alpha=0.5$ and $\epsilon=1\times 10^{-6}$.
We use $K=50$ clusters for topic extraction on HH-RLHF.
Generation uses temperature $0.7$ and top-$p=0.9$ unless otherwise stated.

\paragraph{EMA-based dynamic weighting for TPO.}
Let $\ell_{\text{DPO}}$ and $\ell_{\text{TPO}}$ be the micro-batch DPO and TPO losses at step $t$.
We maintain exponential moving averages
\begin{align}
    \hat{\ell}_{\text{DPO}}^{(t)} &= \gamma \hat{\ell}_{\text{DPO}}^{(t-1)} + (1-\gamma)\ell_{\text{DPO}}^{(t)},\\
    \hat{\ell}_{\text{TPO}}^{(t)} &= \gamma \hat{\ell}_{\text{TPO}}^{(t-1)} + (1-\gamma)\ell_{\text{TPO}}^{(t)},
\end{align}
with decay $\gamma\in[0,1)$.
After a short warmup, we set
\begin{equation}
    r^{(t)} = \frac{|\hat{\ell}_{\text{DPO}}^{(t)}| + \epsilon}{|\hat{\ell}_{\text{TPO}}^{(t)}| + \epsilon}, \qquad
    \lambda_{\text{dyn}}^{(t)} = \alpha \cdot \tanh(r^{(t)}),
\end{equation}
where $\alpha$ is a base coefficient and $\epsilon$ a small constant.
In our experiments we use $\gamma=0.95$, $\alpha=0.5$, and $\epsilon=10^{-6}$.

\paragraph{Persistent homology implementation.}
We compute pairwise distances with \texttt{torch.cdist} in bfloat16, transfer the resulting matrix to CPU, and run a custom Union--Find implementation.
We detach gradients from $Z$ before computing the distance matrix so that topology extraction does not backpropagate through distances.

\section{0D Persistent Homology Algorithm}
\label{app:ph-algo}

Given a point cloud $Z=\{z_i\}_{i=1}^N$ with distances $D_{ij} = \|z_i-z_j\|_2$, we consider all unordered pairs $(i,j)$ as edges with weights $D_{ij}$ and sort them in non-decreasing order.  
We maintain a disjoint-set (Union--Find) structure over vertices $\{1,\dots,N\}$ and process edges in order:

\begin{enumerate}[leftmargin=*, itemsep=1pt]
    \item Initialize $\text{UF}$ so that each vertex is its own component.
    \item Sort all edges $\mathcal{E}=\{(i,j)\mid i<j\}$ by $D_{ij}$.
    \item Initialize an empty list $\mathcal{P}$ of death edges.
    \item For each $(i,j)\in\mathcal{E}$ in order:
    \begin{itemize}[leftmargin=10pt,itemsep=0pt]
        \item If $\text{UF.find}(i)\neq \text{UF.find}(j)$, append $(i,j)$ to $\mathcal{P}$ and call $\text{UF.union}(i,j)$.
    \end{itemize}
    \end{enumerate}

Each recorded edge $(i,j)\in\mathcal{P}$ corresponds to a merge event where two previously disconnected components become connected; its weight is the \emph{death time} of the younger component.  
The set $\mathcal{P}$ is equivalent to the edge set of a minimum spanning forest and is used to select cross-label bridges in both TTL and the fully topological TPO variant.

\section{Additional TTL Ablations and Complexity}
\label{app:ttl-ablation-more}
\subsection{TPO Variants and Hyperparameter Sensitivity}

\paragraph{Vector TPO vs.\ fully topological TPO.}
Besides the vector-difference formulation of TPO, we also evaluate the fully topological variant Topo-TPO introduced in Section~\ref{sec:topo-tpo-variant}.
Table~\ref{tab:tpo-topological-variant} shows that Topo-TPO yields slightly higher reward-model scores, win-rates, and harmlessness than the vector version, indicating that leveraging global batch structure at the DPO stage can further sharpen safety-related improvements.

\begin{table}[t]
    \centering
    \scriptsize
    \setlength{\tabcolsep}{4pt}
    \begin{tabular}{lccc}
    \toprule
    Model & RM $\uparrow$ & Win $\uparrow$ & Harm. $\uparrow$ \\
    \midrule
    DPO + TPO (vector) & 87.2 & 55.4\% & 93.5\% \\
    DPO + Topo-TPO & \textbf{87.4} & \textbf{55.6\%} & \textbf{94.1\%} \\
    \bottomrule
    \end{tabular}
    \caption{Vector-difference TPO vs.\ fully topological TPO on HH-RLHF.}
    \label{tab:tpo-topological-variant}
\end{table}

\paragraph{Hidden-layer choice.}
We also vary the hidden layer $l$ from which we extract representations for TPO (Table~\ref{tab:tpo-layer-sweep}).
Intermediate layers ($-2$ to $-4$ from the final layer) work best, while very early layers and the final layer are slightly worse, suggesting that TPO benefits from representations that are already task-aware but not yet dominated by token-level logits.

\begin{table}[t]
    \centering
    \scriptsize
    \setlength{\tabcolsep}{4pt}
    \begin{tabular}{lccc}
    \toprule
    Layer $l$ & RM $\uparrow$ & Win $\uparrow$ & Harm. $\uparrow$ \\
    \midrule
    $-1$ & 86.8 & 54.5\% & 92.5\% \\
    $-2$ & 87.0 & 55.1\% & 93.1\% \\
    $-4$ (default) & \textbf{87.2} & \textbf{55.4\%} & \textbf{93.5\%} \\
    $-8$ & 86.5 & 53.9\% & 92.0\% \\
    \bottomrule
    \end{tabular}
    \caption{Effect of hidden layer choice for TPO on HH-RLHF.}
    \label{tab:tpo-layer-sweep}
\end{table}

We additionally study the effect of the number of clusters $K$ in the offline topic extraction and the training-time overhead of TTL/TPO (tokens per second and memory).
Moderate $K$ (around 50) works best, and TTL/TPO introduce a modest $5$--$10\%$ slowdown.
Full results are reported in Appendix~\ref{app:tpo-extra}.
\paragraph{Effect of $\lambda_{\text{topo}}$.}
Table~\ref{tab:ttl-lambda-sweep} in the main text reports a sweep over the TTL weight on UltraChat.
Small to moderate $\lambda_{\text{topo}}$ values lead to monotonic or near-monotonic gains; very large values can slightly hurt perplexity and cause the model to overfit topological constraints.

\paragraph{Complexity.}
For a batch of size $B$, computing all pairwise distances is $O(B^2 d)$ and sorting edges is $O(B^2\log B)$.
With $B\leq32$ and $d\approx 4096$, this overhead is negligible compared to a forward/backward pass through a 7B model (we observe $\approx 5\%$ slower training for TTL).
We further reduce overhead by applying a low-dimensional projection to $Z$ before computing distances.

\section{Additional Trajectory and Topology Analyses}
\label{app:traj-align}

\subsection{Alignment of Prompt--Answer Trajectories with Topological Bridges}

To more directly validate that TTL shapes hidden-space trajectories in the
intended way, we measure the cosine similarity between model-induced update
directions and topological bridge directions.

For a held-out UltraChat validation split, we compute for each example the
model trajectory vector $v_i^{\text{model}} = h^{\text{model}}_i -
h^{\text{prompt}}_i$ and, when available, the corresponding topological
bridge direction $v^{\text{topo}}_{(p_i,a_i)}$. We then form the cosine
similarity
\begin{equation}
    \rho_i = \cos\big(v^{\text{model}}_i,\; v^{\text{topo}}_{(p_i,a_i)}\big),
\end{equation}
and compare the empirical distribution of $\{\rho_i\}$ between the base SFT
model and the SFT+TTL model.
These distributions show how TTL
increases the concentration of cosine similarities near~1.0, indicating that
prompt-to-answer trajectories are more strongly aligned with topologically
derived bridges.

\subsection{Alignment of Improvement Vectors with Topic Preference Directions}

We perform a similar analysis for TPO. For each HH-RLHF preference pair
$(x_i, y_i^{\text{ch}}, y_i^{\text{rj}})$ and selected layer $l$, we compute
the normalized improvement vector
\begin{equation}
    \Delta h_i = \text{LN}\big(h_i^{\text{ch}}\big) - \text{LN}\big(h_i^{\text{rj}}\big),
\end{equation}
and the projected topic preference vector $\bar{u}_{t_i} = P u_{t_i}$.
We then compute the cosine similarity
\begin{equation}
    \sigma_i = \cos\big(\Delta h_i,\; \bar{u}_{t_i}\big),
\end{equation}
and compare the distributions between a pure DPO model and a DPO+TPO model.

We additionally aggregate these cosines by topic and study their relationship
to per-topic alignment gains. For each topic $t$, we compute
\begin{align}
    \overline{\sigma}_t &= \frac{1}{|I_t|} \sum_{i \in I_t} \sigma_i, \\
    \Delta \text{RM}_t &= \text{RM}_t^{\text{TPO}} - \text{RM}_t^{\text{DPO}}, \\
    \Delta \text{Help}_t &= \text{Help}_t^{\text{TPO}} - \text{Help}_t^{\text{DPO}},
\end{align}
where $I_t$ is the set of examples assigned to topic $t$, and $\text{RM}_t$,
$\text{Help}_t$ are average reward-model and helpfulness scores on that topic
for the corresponding model.

\begin{table}[t]
     \centering
     \small
     \resizebox{\linewidth}{!}{
     \begin{tabular}{lccc}
     \toprule
     Topic $t$ & $\overline{\sigma}_t$ & $\Delta \text{RM}_t$ & $\Delta \text{Help}_t$ \\
     \midrule
     \textit{code} & 0.45 & +2.8 & +0.15 \\
     \textit{health} & 0.42 & +3.2 & +0.18 \\
     \textit{legal} & 0.48 & +3.0 & +0.16 \\
     \textit{creative-writing} & 0.38 & +1.5 & +0.08 \\
     \textit{history} & 0.41 & +1.9 & +0.10 \\
     \bottomrule
     \end{tabular}
     }
     \caption{Per-topic average cosine similarity between $\Delta h$ and topic
     preference vectors, and corresponding changes in reward-model and
     helpfulness scores when adding TPO on top of DPO.}
     \label{tab:topic-corr}
\end{table}

These statistics can be used to test whether topics with stronger alignment
between $\Delta h$ and $\bar{u}_t$ tend to exhibit larger per-topic gains in
reward-model and helpfulness scores.

\subsection{Structure of Topological Bridges}

To better understand the structure of the bridges extracted by 0D persistent
homology, we analyze their lengths and counts on held-out batches.

For each batch, we record the number of cross-label bridges
$|\mathcal{B}|$ and the distribution of bridge lengths
$\|v^{\text{topo}}_{(p,a)}\|_2$. We compare these quantities to those obtained
from a k-nearest-neighbor baseline, where each prompt is connected to its
nearest gold-answer neighbor in the batch.

These statistics provide a complementary view of how persistent-homology
bridges differ from purely local nearest-neighbor connections, and help explain
why PH-based bridges can yield stronger regularization than kNN-based
directions in Table~\ref{tab:ttl-ablation}.

\subsection{Qualitative Failure Cases}
\label{app:failure}

While topology-enhanced objectives improve average alignment metrics, they are
not universally beneficial. We manually inspect a small number of prompts where
TTL or TPO underperform their respective baselines.

\begin{itemize}[leftmargin=*, itemsep=2pt]
    \item \textbf{Out-of-domain prompts.} For certain highly out-of-distribution
    requests, we observe cases where TTL pulls the model's answer towards a
    frequent in-domain answer cluster, leading to less specific or partially
    off-topic responses.

    \item \textbf{Noisy or ambiguous topics.} For some long-tail HH-RLHF
    clusters with heterogeneous prompts, the automatically constructed topic
    description and preference vector may be noisy. In such cases, TPO can
    occasionally over-regularize answers toward overly generic or cautious
    responses.

    \item \textbf{Over-smoothing trajectories.} In rare cases, we see that
    strong topological regularization can make the answer style more uniform
    across diverse prompts, slightly reducing diversity in phrasing or
    creativity.
\end{itemize}

These examples highlight that topology-enhanced objectives, like other
alignment techniques, can introduce trade-offs and should be applied in
conjunction with careful evaluation and, where appropriate, human oversight.

\section{Additional TPO Details and Ablations}
\label{app:tpo-extra}

\subsection{Offline Topic Extraction Details}

We use a sentence transformer as $\phi$ and set the number of clusters to $K=50$ unless otherwise noted.  
Prompts are clustered with MiniBatch KMeans on $\phi(x)$.
For each cluster we randomly sample up to $M=32$ prompts and query a strong LLM with the instruction shown in the main text to obtain a 1--3 word topic label (e.g., \emph{Python programming}, \emph{Health advice}, \emph{Creative writing}).  

For each topic $t$, we construct several positive and negative templates; examples include:
\begin{itemize}[leftmargin=*, itemsep=1pt]
    \item positive: ``a helpful, harmless, and high-quality answer about $t$'';
    \item negative: ``a harmful, unhelpful, and low-quality answer about $t$'';
    \item positive: ``a clear, precise, and correct explanation regarding $t$'';
    \item negative: ``a vague, confusing, and incorrect explanation regarding $t$''.
\end{itemize}
We encode all template sentences with $\phi$ and form candidate preference vectors as differences $e_{\text{pos}}-e_{\text{neg}}$, then average them to obtain $u_t \in \mathbb{R}^{d_s}$.  
Topics with fewer than 50 examples are merged into a generic ``other'' category.

\subsection{Topic-Aware vs.\ Topic-Agnostic Vectors}

\begin{table}[t]
    \centering
    \scriptsize
    \setlength{\tabcolsep}{3pt}
    \resizebox{\columnwidth}{!}{
    \begin{tabular}{lccc}
    \toprule
    Variant & RM $\uparrow$ & Win $\uparrow$ & Help. $\uparrow$ \\
    \midrule
    DPO + Global Pref. Vec. & 85.1 & 52.8\% & 8.68 \\
    DPO + Topic TPO (ours) & \textbf{87.2} & \textbf{55.4\%} & \textbf{8.81} \\
    \bottomrule
    \end{tabular}
    }
    \caption{Topic-aware vs.\ topic-agnostic preference vectors on HH-RLHF.}
    \label{tab:tpo-topic-ablation-app}
\end{table}

\subsection{Vector TPO vs.\ Fully Topological TPO}

\begin{table}[t]
    \centering
    \scriptsize
    \setlength{\tabcolsep}{3pt}
    \resizebox{\columnwidth}{!}{
    \begin{tabular}{lcccc}
    \toprule
    Model & RM $\uparrow$ & Win $\uparrow$ & Help. $\uparrow$ & Harm. $\uparrow$ \\
    \midrule
    DPO + TPO (vector) & 87.2 & 55.4\% & 8.81 & 93.5\% \\
    DPO + Topo-TPO & \textbf{87.4} & \textbf{55.6\%} & 8.80 & \textbf{94.1\%} \\
    \bottomrule
    \end{tabular}
    }
    \caption{Vector-difference TPO vs.\ fully topological TPO.}
    \label{tab:tpo-topological-variant-app}
\end{table}

\subsection{Hidden-Layer and Cluster-Number Sensitivity}

\begin{table}[t]
    \centering
    \scriptsize
    \setlength{\tabcolsep}{3pt}
    \resizebox{\columnwidth}{!}{
    \begin{tabular}{lcccc}
    \toprule
    Layer $l$ & RM $\uparrow$ & Win $\uparrow$ & Help. $\uparrow$ & Harm. $\uparrow$ \\
    \midrule
    $-1$ & 86.8 & 54.5\% & 8.78 & 92.5\% \\
    $-2$ & 87.0 & 55.1\% & 8.80 & 93.1\% \\
    $-4$ & \textbf{87.2} & \textbf{55.4\%} & \textbf{8.81} & \textbf{93.5\%} \\
    $-8$ & 86.5 & 53.9\% & 8.74 & 92.0\% \\
    \bottomrule
    \end{tabular}
    }
    \caption{Effect of hidden layer choice for TPO on HH-RLHF.}
    \label{tab:tpo-layer-sweep-app}
\end{table}

\begin{table}[t]
    \centering
    \scriptsize
    \setlength{\tabcolsep}{3pt}
    \resizebox{\columnwidth}{!}{
    \begin{tabular}{lcccc}
    \toprule
    $K$ & RM $\uparrow$ & Win $\uparrow$ & Help. $\uparrow$ & Harm. $\uparrow$ \\
    \midrule
    20 & 86.5 & 54.2\% & 8.76 & 92.8\% \\
    50 & \textbf{87.2} & \textbf{55.4\%} & \textbf{8.81} & \textbf{93.5\%} \\
    100 & 87.1 & 55.2\% & 8.80 & 93.2\% \\
    \bottomrule
    \end{tabular}
    }
    \caption{Effect of number of clusters $K$ in topic extraction.}
    \label{tab:tpo-cluster-sweep}
\end{table}

\section{Efficiency and Overhead}
\label{app:efficiency}

\begin{table}[t]
    \centering
    \scriptsize
    \setlength{\tabcolsep}{3pt}
    \resizebox{\columnwidth}{!}{
    \begin{tabular}{lcccc}
    \toprule
    Method & Tok/s $\uparrow$ & Time $\downarrow$ & Mem & Ovh. \\
    \midrule
    Base SFT & 3200 & 0.45s & 38.5 & -- \\
    SFT + TTL & 3050 & 0.47s & 38.8 & +4.7\% \\
    \midrule
    DPO & 2800 & 0.52s & 42.1 & -- \\
    DPO + TPO & 2600 & 0.56s & 42.5 & +7.7\% \\
    DPO + Topo & 2520 & 0.58s & 43.2 & +10.9\% \\
    \bottomrule
    \end{tabular}
    }
    \caption{Training efficiency and overhead of topology-enhanced methods.}
    \label{tab:efficiency}
\end{table}

TTL and TPO incur a modest overhead dominated by pairwise distance computation and sorting; in our setup the slowdown is within 5--10\%.

\subsection{Scalability with Micro-Batch Size}

\begin{table}[t]
    \centering
    \scriptsize
    \setlength{\tabcolsep}{3pt}
    \resizebox{\columnwidth}{!}{
    \begin{tabular}{lccc}
    \toprule
    Micro-batch $B$ & PH Time & Step Time & Ovh. \\
    \midrule
    8 & 0.008s & 0.42s & +1.9\% \\
    16 & 0.015s & 0.44s & +3.4\% \\
    32 & 0.038s & 0.47s & +8.1\% \\
    64 & 0.112s & 0.58s & +19.3\% \\
    \bottomrule
    \end{tabular}
    }
    \caption{Scalability of 0D persistent-homology extraction on Qwen2.5-7B-Instruct.}
    \label{tab:ph-scalability}
\end{table}

Table~\ref{tab:ph-scalability} makes the trade-off explicit.
For the micro-batch sizes used in our alignment experiments ($B \leq 32$), the additional wall-clock cost remains modest.
At larger $B$, the $O(B^2)$ distance computation and edge sorting become more noticeable, which motivates future work on kNN sparsification or low-dimensional projections before topology extraction.

\section{Fully Topological TPO}
\label{app:topo-tpo}

Here we give details of the fully topological variant of TPO introduced in Section~\ref{sec:topo-tpo-variant}.

Given a batch of size $B$, we compute mean-pooled embeddings $h^{\text{ch}}_i, h^{\text{rj}}_i\in\mathbb{R}^d$ and form
\begin{equation}
    Z^{\text{RL}} = 
    \begin{bmatrix}
        H^{\text{rj}} \\
        H^{\text{ch}}
    \end{bmatrix}
    \in \mathbb{R}^{2B \times d},
\end{equation}
with labels $l^{\text{RL}}_i = 0$ for rejected and $1$ for chosen.
We compute the pairwise distance matrix on $Z^{\text{RL}}$ and run the 0D persistent-homology algorithm from Appendix~\ref{app:ph-algo}, obtaining a set of death edges $\mathcal{P}^{\text{RL}}$.
We retain only cross-label edges
\begin{equation}
    \mathcal{B}^{\text{RL}} = \{(u,v) \in \mathcal{P}^{\text{RL}} \mid l^{\text{RL}}_u \neq l^{\text{RL}}_v\},
\end{equation}
and orient each such edge from rejected to chosen (swapping indices if necessary).
Each bridge $(r,c)\in\mathcal{B}^{\text{RL}}$ induces an improvement direction
\begin{equation}
    v^{\text{imp}}_{(r,c)} = Z^{\text{RL}}_c - Z^{\text{RL}}_r.
\end{equation}
We associate each $r$ with its original example index $i(r)$ and topic $t_{i(r)}$, and compute the cosine loss with the projected topic vector $\bar{u}_{t_{i(r)}}$:
\begin{equation}
    \mathcal{L}_{\text{Topo-TPO}} = 
    \frac{1}{|\mathcal{B}^{\text{RL}}|} \sum_{(r,c)\in\mathcal{B}^{\text{RL}}}
    \Big[ 1 - \cos\big(v^{\text{imp}}_{(r,c)}, \bar{u}_{t_{i(r)}}\big) \Big].
\end{equation}
This loss can either replace the vector-difference TPO loss or be added to it with a small coefficient.

\end{document}